# An Algorithm to perform Covariance-Adjusted Support Vector Classification in Non-Euclidean Spaces

Satyajeet Sahoo[1], and Jhareswar Maiti[1,2]

[1]Department of Industrial Engineering, Indian Institute of Technology Kharagpur, Kharagpur-721302, West Bengal, India
[2]CoE in Safety Engineering and Analytics, Indian Institute of Technology Kharagpur, Kharagpur-721302, West Bengal, India

**Abstract. Traditional Support Vector Machine (SVM) classification is carried out by finding the max-margin classifier for the training data that divides the margin space into two equal sub-spaces. This study demonstrates limitations of performing Support Vector Classification in non-Euclidean spaces by establishing that the underlying principle of max-margin classification and Karush Kuhn Tucker (KKT) boundary conditions are optimal only in the Euclidean vector spaces. The study establishes a methodology to perform Support Vector Classification in Non-Euclidean Spaces by incorporating data covariance into the optimization problem using Cholesky Decomposition of respective class covariance structure. It also demonstrates that in non-Euclidean spaces KKT modelling is sub-optimal as the principle of maximum margin is a function of intra-class data covariances and the classifier obtained separates the margin space in ratio of the respective class population covariance matrix. The study proposes an algorithm to iteratively estimate the population covariance-adjusted SVM classifier in non-Euclidean space from sample covariance matrices of the training data. The effectiveness of this SVM classification approach is demonstrated by applying the classifier on multiple datasets and comparing the performance with traditional SVM kernels and whitening algorithms. The Cholesky-SVM model shows marked improvement in the accuracy, precision, F1 scores and ROC performance compared to linear and other kernel SVMs.**

## 1 Introduction

Support Vector Machines (SVM), derived from Vapnik's statistical learning theory [1] is a powerful kernel-based machine learning tool that is suitable for both classification and regression tasks, and hence is widely used in diverse fields ranging from pattern recognition [2] to text classification [3], image classification [4] and forecasting in finance [5]. Unlike other classification algorithms that focus on empirical risk minimization, SVM focuses on structural risk minimization [6] [7]. It achieves this by identifying a separating linear hyperplane with maximum margin from the margin-edge hyperplanes that bound the two data classes in the N-dimensional space, N being the number of features. The parameters of the margin-edge hyperplanes and the decision boundary are estimated by constructing an optimization problem and solving using quadratic programming. When the data is not linearly separable, kernel tricks are applied where kernels transform data from input space into a linearly-separable higher dimensional feature space [8].

The equation of the linear hyperplanes for both the decision boundary and margin boundaries, and the margin calculations are derived using Euclidean distance formula assuming that the space is Euclidean. However, P.C. Mahalanobis [9] showed that the input space (also called the statistical space or sample space) is Non-Euclidean, and the statistical distance which incorporates the covariance matrix into the Euclidean distance formula is the true measure of distance between two data points in the statistical space. Hence calculating the SVM decision boundary and the margins based on the Euclidean distance in the input space/statistical space carries risk of misclassification. Secondly, the concept of Max-Margin Classification considers the decision boundary to be equidistant from either margin hyperplane. SVM optimization problem uses KKT boundary conditions, where only the support vectors at the margin boundaries play a role in deciding the decision boundary. The role of the other data points, as well as information about the data class distributions and inherent variance-covariance structure of the data plays a minimal role [10]. However, a separating hyperplane should divide the margin space in ratio of the data dispersion of each class- a higher margin should be provided for the data class having higher dispersion (higher variance) and lower margins for the data class having lower dispersion (lower variance) as it is more cohesive and compact. The entire dataset (through variance) should influence the calculation of decision boundary. Hence, adjustments need to be made for class distribution variance while calculating and maximizing the margins in the optimization problem.

Several studies have acknowledged the aforementioned issue and have recommended various approaches to incorporate variance into the SVM optimization problem. Tsang et al [11] incorporated covariance information in one-class SVMs by using the Mahalanobis distance instead of Euclidean distance to calculate the margin. Peng and Xu [12] incorporated Mahalanobis distance into twin support vector machine (TSVM) to determine two optimization problems to determine the two nonparallel separating hyperplanes. Ke et al [13] presented a Mahalanobis distance based biased least squares support vector machine (MD-BLSSVM) to classify PU data. Huang et. al [14] proposed the maxi-min margin machine that incorporates class distribution information into decision boundary optimization problem using statistical distance. Wang et. al. [15] proposed weighted Mahalanobis distance Kernels for SVMs that incorporates covariance information into existing kernels. Zafeiriou et. al. [9] proposed the minimum class variance SVMs (MCVSVMs) by optimizing Fisher's Discriminant Analysis where a within-class scatter matrix is incorporated in optimization problem to account for data variance. Analysis of the optimization problems formulated in those studies led us to identify certain gaps, particularly in formulation of the constraint equations. In this study we have tried to rectify those gaps by proposing Covariance-Adjusted Support Vector Machine (CSVM)-building upon the concepts that (1) statistical space is different from the Euclidean space, (2) SVM is optimal in Euclidean space only as it is derived using Euclidean distance, and (3) Mahalanobis distance is essentially a transformation of the data from the statistical space to the Euclidean space ( Sahoo & Maiti (2025)). Using these concepts, we first transform data from Non-Euclidean statistical space to Euclidean space by performing Cholesky Decomposition of intra-class covariance matrix for each data class, and using the lower triangular matrix as transformation matrix. Then we formulate the SVM optimization problem in the transformed Euclidean space. The optimization problem thus formulated in the transformed space is not only derived from first-principles, but is also consistent dimension and vector space wise. Here, the Cholesky decomposed lower-triangular matrix of the covariance matrix transforms the data from statistical space to Euclidean space, thus mirroring the kernel trick. From the covariance adjusted optimization problem formulation, we can prove that (1) an N-class support-vector classification problem results in N decision boundaries in the input space; (2) equivalence between statistical space and the Euclidean space implies that the decision boundary should split the margin between the margin boundaries in the input space in ratio of respective data class covariances; (3) KKT Boundary conditions are not optimal while attempting SVM in a Non-Euclidean Space. We address the limitations of calculating population covariance matrix in absence of test data label information by proposing the SM Algorithm- that iteratively calculates the population covariance matrix and performs Covariance-adjusted SVM (CSVM) from training data information. Finally, effectiveness of CSVM is demonstrated by applying on 5 standard datasets and comparing the results with traditional linear and other kernel SVMs, as well as with other prevalent whitening methods. CSVM shows better classification table metrics and higher AUC values in ROC curves compared to other models.

The remainder of this paper is organized as follows: Section 2 briefly describes the background of SVM and the limitations herewith. Section 3 presents a mathematical derivation of the vector space transformation from statistical to Euclidean using Cholesky decomposition and formulates the optimization problem of the SVM classifier after adjusting for data covariance in the Euclidean space. Section 4 gives the SM Algorithm to iterately calculate population covariance from training data sample covariance. Comparison with other studies done in literature and other whitening approaches is done in section 5. Section 5 gives the results of application of CSVM and other SVM approaches on 5 datasets and performance is compared using precision, recall, accuracy, F1 scores and ROC curves. Finally, sections 6 concludes with a summary of the contributions, limitations and future scope of work.

### 1.1 Support Vector Machine

Support Vector Machine (SVM) is a supervised learning model that identifies a separating hyperplane that acts as a decision boundary and classifies unlabeled data into one of the two labels $y \in \{+1, -1\}$. To estimate the parameters of the hyperplane, an anointed training dataset is used, consisting of N observations. For each observation, values of some pre-identified variables are collected and stored as feature vectors, along with its manually anointed label depicting which class it belongs to. Hence each observation is represented as a point in the k-dimensional space $R^k$ represented by ($x_i$,$y_i$) where $x_i$ is the feature vector and $y_i$ is the label.

In linear SVM, the separating hyperplane is represented by the equation:

$$\Theta^T x + \Theta_0 = 0 \quad (1)$$

Where $\theta \in R^k$ is the weight vector (normal to the hyperplane), x is the feature vector and $\theta_0$ is the bias. The objective is to estimate the values of $\theta$ and $\theta_0$ to obtain the resulting hyperplane with the largest margin while minimizing hinge loss. This can be obtained by solving the quadratic programming problem:

$$\text{Min } \frac{1}{2}\theta^T\Theta + \lambda\sum_{i=1}^{N}\xi_i \quad (2)$$
$$\text{s.t. } y_i(\theta^T x_i + \theta_0) \geq 1\text{-}\xi_i,\ \xi_i \geq 0 \quad \forall \text{i} = 1,2,\dots\text{N} \quad (3)$$

Where $\lambda$ is the parameter of tradeoff between regularization and hinge loss, and $\xi_i$ are the slack variables for each ($x_i$,$y_i$).

Solving its dual problem obtains the appropriate soft margin hyperplane. First the Lagrangian function for the optimization problem is obtained which is then converted to dual form and the parameters estimated using quadratic programming. The separating hyperplane is estimated as

$$\text{f(x)} = \text{sgn}\ (\theta^T x + \theta_0) = \text{sgn}\ (\sum_{i=1}^{N}\alpha_i y_i x_i^T x + \theta_0) \quad (4)$$

where $\alpha_i$ is the respective Lagrangian multiplier of the respective feature vector $x_i$. The training observations lying on the margin hyperplanes have the value of the Lagrangian $\alpha_i$ >0 and the rest as 0 (KKT Condition). Hence the decision function is exclusively determined by the margin vectors, also known as support vectors.

One of the limitations of SVM is that it depends exclusively on the support vectors without accounting for any other characteristics of the data. For example, consider the following two classes as shown in Figure 1:

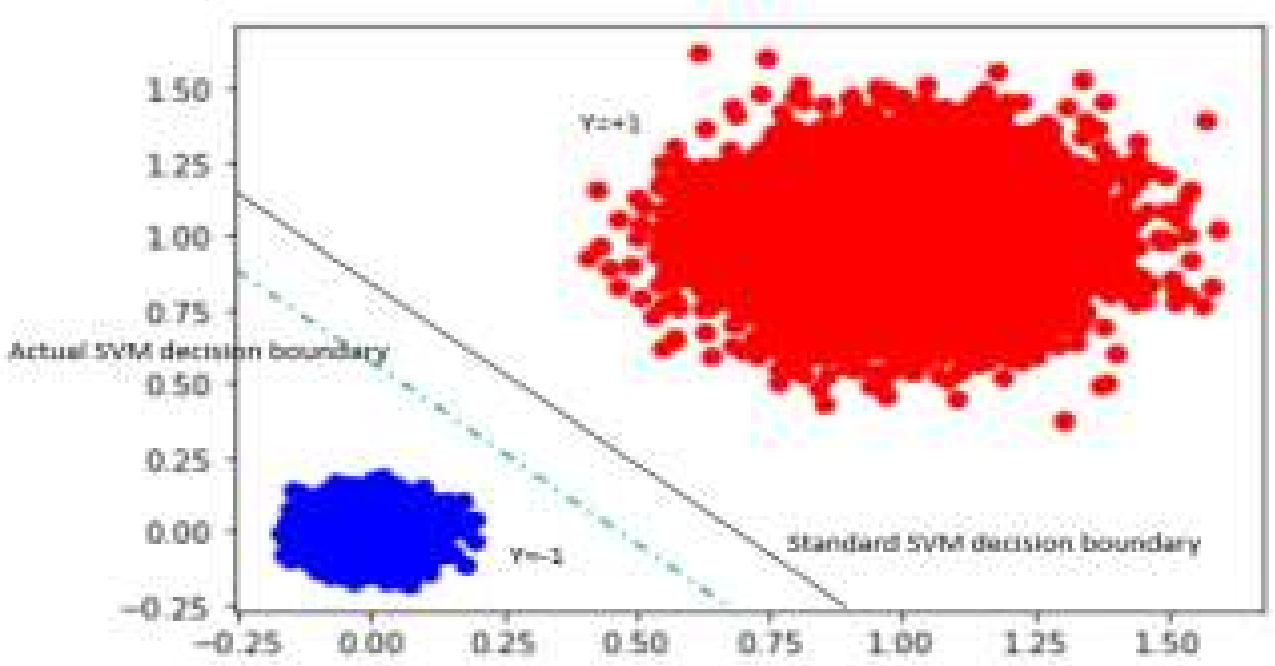

Figure 1: Decision boundaries for data with different variances

In standard linear SVM, the separating hyperplane has equal margins from the support vectors of both the classes. The margin is $\frac{1}{\|\theta\|}$ from each margin hyperplane. However, the data with class label y=+1 has higher dispersion/scatter compared to data with class label y=-1. Dispersion in data is represented by its variance. Hence it makes sense that rather than having equal margins from both the classes, a separating hyperplane should leave higher margin for the data class having higher dispersion (higher variance) and lower margins for the data class having lower dispersion (lower variance).

To address these limitations, we propose Covariance-Adjusted Support Vector Machine (CSVM) data undergoes vector space transformation using Cholesky factorization of covariance matrix and support vector optimization problem is formulated and solved in the transformed Euclidean space.

## 2 Vector space transformation to Euclidean space for SVM

**Definition 2.1.** ***The classical Support Vector Machine assumes an identity covariance matrix. While the Karush-Kuhn-Tucker (KKT) conditions provide a unique global L2-norm based optimal solution, this geometric margin is statistically sub-optimal when data exhibits significant class-specific non-identical covariance structures ($\Sigma_1 \neq \Sigma_2 \neq I$). In such "Non-Euclidean" statistical manifolds, the KKT conditions optimize for a geometric separation that does not coincide with the Bayes optimal decision boundary.***

**Proof:** For a probability distribution of a population, distance between a point and the mean in statistical/input space given by the Mahalanobis distance written as:

$$(X - \mu)^T \Sigma^{-1} (X - \mu) \quad (5)$$

As per Sahoo and Maiti [16], Mahalanobis distance can be written as:

$$(X-\mu)^T\Sigma^{-1}(X-\mu) = (X-\mu)^T(\Psi\Psi^T)^{-1}(X-\mu)$$

$$= (X-\mu)^T(\Psi^T)^{-1}(\Psi)^{-1}(X-\mu) = (X-\mu)^T(\Psi^{-1})^T(\Psi)^{-1}(X-\mu)$$

$$= \left[\left(\Psi^{-1}(X-\mu)\right)\right]^T\left[\left(\Psi^{-1}(X-\mu)\right)\right] \quad (6)$$

Where Σ is the population covariance matrix and

$$\Sigma = \Psi\Psi^T \quad (7)$$

is the Cholesky decomposition of Σ resulting in the lower triangular matrix Ψ.

While the LHS in (5) represents distance between raw data and the data mean in statistical/input space, its equivalence in the RHS is the distance between the data point and the mean that are both transformed from the statistical space to a new vector space using $\Psi^{-1}$ , and represented by the inner product $[\Psi^{-1}(X-\mu)]^T[\Psi^{-1}(X-\mu)]$. The new vector space has following properties:

- The space has an inner product, hence it is an inner product space.
- The inner product represents the distance between the points.
- If $\mu = 0$, the inner product represents the norm.

All these are properties of Euclidean space. Hence the new vector space where the original data was transformed using $\Psi^{-1}$ is the Euclidean Space with the transformed data is given by

$$X^{\text{Euclidean}} = \Psi^{-1}X^{\text{Input}} \quad (8)$$

and the original statistical/input space is a non-Euclidean space.

Mahalanobis transformation uses the covariance matrix of the probability distribution, which is derived from the population/sample. In SVM of binary classification, the dataset belongs to two classes given by y=1 or y=-1. As these are two distinct labels with distinct features and parameters, they can be considered as two distinct populations, which correspond to two distinct distributions. These distributions have their unique covariance structure, hence will be characterized by their distinct covariance matrix $\Sigma_{y=1}$ and $\Sigma_{y=-1}$ respectively. Accordingly, the data transformation can be given by

$$X_{y=1}^{\text{Euclidean}} = \Psi_{y=1}^{-1}X_{y=1}^{Input} \quad (9)$$

$$X_{y=-1}^{\text{Euclidean}} = \Psi_{y=-1}^{-1}X_{y=-1}^{Input} \quad (10)$$

In SVM the objective is to find out a linear hyperplane that separates the data with maximum margin from the data points of either class. The equation of the hyperplane and the magnitude of the margin derived from the Euclidean distance formula and principles of cartesian coordinate system. Since these operations are valid in the Euclidean space only, the following lemma is proposed:

***Lemma 3.1. Principles of support vector classification (calculation of the maximum margin and identification of maximum margin hyperplane) are optimal only when the data is transformed from the input/statistical space to the Euclidean space using $\boldsymbol{\Psi^{-1}}$, which is inverse of the lower triangular matrix obtained by Cholesky Decomposition of the population covariance matrix. The SVM optimization problem, loss function and constraints are formulated in the transformed space.***

*Proof:* In the Euclidean space, the equation of the maximum margin classifier becomes (considering hard margin SVM, $\xi_i$=0)

$$\Theta^{T}X^{\text{Euclidean}} + \Theta_0 = 0 \tag{11}$$

And the margin is given by

$$\frac{1}{\sqrt{\Theta^{T}\Theta}} \tag{12}$$

So the optimization problem becomes

$$\text{Min}\ \frac{1}{2}\Theta^{T}\ \Theta \tag{13}$$

Subject to

$$y_i(\Theta^{T}X^{\text{Euclidean}} + \Theta_0) \geq 1 \tag{14}$$

Formulating Euclidean space-non Euclidean input space equivalence, the decision boundary given in (9) becomes (considering hard margin SVM, $\xi_i$=0)

$$\Theta^{T}\Psi_{y=1}^{-1}X_{y=1}^{\text{Input}} + \Theta_0 = 0 \quad \text{or} \quad \left(\left(\Psi_{y=1}^{-1}\right)^{T}\Theta\right)^{T} X_{y=1}^{\text{Input}} + \Theta_0 = 0 \tag{15}$$

for population of data points labelled y=1.

What is the margin of the linear classifier hyperplane given in (10) in the input space? If hypothetically, we apply the rules of Cartesian coordinate geometry and try to calculate the margin from the margin boundary to the classifier, the margin is

$$\frac{1}{\sqrt{\left(\left(\Psi_{y=1}^{-1}\right)^{T}\theta\right)^{T}\left(\left(\Psi_{y=1}^{-1}\right)^{T}\theta\right)}} = \frac{1}{\sqrt{\theta^{T}\left(\left(\Psi_{y=1}^{-1}\right)^{T}\right)^{T}\left(\left(\Psi_{y=1}^{T}\right)^{-1}\theta\right)}} = \frac{1}{\sqrt{\theta^{T}\Psi_{y=1}^{-1}\left(\Psi_{y=1}^{T}\right)^{-1}\theta}} =$$

$$\frac{1}{\sqrt{\theta^{T}\left(\Psi_{y=1}\right)^{-1}\left(\Psi_{y=1}^{T}\right)^{-1}\theta}} = \frac{1}{\sqrt{\theta^{T}\left(\Psi_{y=1}^{T}\Psi_{y=1}\right)^{-1}\theta}} = \frac{1}{\sqrt{\theta^{T}\left(\Psi_{y=1}^{-1}\Psi_{y=1}\Psi_{y=1}^{T}\Psi_{y=1}\right)^{-1}\theta}} =$$

$$\frac{1}{\sqrt{\theta^{T}\left(\Psi_{y=1}^{-1}\Sigma_{y=1}\Psi_{y=1}\right)^{-1}\theta}} = \frac{1}{\sqrt{\theta^{T}\Psi_{y=1}^{-1}\Sigma_{y=1}^{-1}\Psi_{y=1}\theta}} \quad (16)$$

Hence the margin maximization problem in the input space for y=1 becomes

$$\text{Maximize } \frac{1}{\theta^{T}\Psi_{y=1}^{-1}\Sigma_{y=1}^{-1}\Psi_{y=1}\theta} \quad (17)$$

Satisfying the constraint

$$\theta^{T}\Psi_{y=1}^{-1}X_{y=1}^{Input} + \theta_0 \geq 1 \quad (18)$$

Similarly for the data label y=-1, the margin maximization problem becomes

$$\text{Maximize } \frac{1}{\theta^{T}\Psi_{y=-1}^{-1}\Sigma_{y=-1}^{-1}\Psi_{y=-1}\theta} \quad (19)$$

Satisfying the constraint

$$\theta^{T}\Psi_{y=-1}^{-1}X_{y=-1}^{Input} + \theta_0 \leq -1 \quad (20)$$

There are two sets of optimization problems to be solved. This gives rise to the following lemma:

***Corollary 3.1. For a two-class problem, the application of SVM in the input space domain generates not one, but two unique optimization problem formulations resulting in two unique linear classifiers—each input space having its own linear classifier. In an N-class problem, there will be N data class distributions and N input spaces; hence, there will be N linear classifiers.***

Comparison of maximized margins obtained from equations (14) between classifiers of y=1 and y=-1 result in

$$\frac{\text{Margin}_{y=1}}{\text{Margin}_{y=-1}} = \sqrt{\frac{\theta^{T}\Psi_{y=-1}^{-1}\Sigma_{y=-1}^{-1}\Psi_{y=-1}\theta}{\theta^{T}\Psi_{y=1}^{-1}\Sigma_{y=1}^{-1}\Psi_{y=1}\theta}} = f\left(\frac{\theta^{T}\Psi_{y=-1}^{-1}\Sigma_{y=-1}^{-1}\Psi_{y=-1}\theta}{\theta^{T}\Psi_{y=1}^{-1}\Sigma_{y=1}^{-1}\Psi_{y=1}\theta}\right) = f\left(\frac{\Sigma_{y=-1}^{-1}}{\Sigma_{y=1}^{-1}}\right) \quad (21)$$

The following lemma is proposed:

***Corollary 3.2. In the input space, margin for each data class is a function of respective population covariance matrix, $\frac{1}{\Sigma^{-1}}$. Hence the assumptions of KKT boundary conditions are not optimal in the input space as each data point (other than support vectors) contributes to Σ.***

Hence unlike the original support vector machine algorithm, in a Non-Euclidean space the margin of the classifier is dependent on the intra-class data covariance (variance is essentially covariance of a random variable with itself).

# 3 Algorithm for SVM Classification from sample covariance information

CSVM vector transformation requires knowing the population covariance matrix Σ, which is not possible as the test data labels are unknown. Here we propose an algorithm here to perform Covariance-Adjusted SVM classification using the sample covariance matrices Sy=1 and Sy=−1 that can be calculated from training data (We call it the SM Algorithm). We have established that if we do SVM in the Euclidean space, the corresponding linear classifier in the input space divides the margin as a function of $\frac{1}{\Sigma^{-1}}$ as given in equation (14), and it can give an indicative location of the test data w.r.t the classifier and its labels. Hence in this algorithm we iteratively perform the following steps: (1) Calculate the covariance matrices for the available labelled data (starting with training data) to obtain $S_{y=-1}$ and $S_{y=-1}$; (2) Find the SVC linear classifier that divides the margin in the ratio $f\left(\frac{\theta^T \Psi_{y=-1}^{-1} \Sigma_{y=-1}^{-1} \Psi_{y=-1} \theta}{\theta^T \Psi_{y=1}^{-1} \Sigma_{y=1}^{-1} \Psi_{y=1} \theta}\right)$ i.e. $\sqrt{\frac{\theta^T \Psi_{y=-1}^{-1} \Sigma_{y=-1}^{-1} \Psi_{y=-1} \theta}{\theta^T \Psi_{y=1}^{-1} \Sigma_{y=1}^{-1} \Psi_{y=1} \theta}}$, and list all the data points on either side of the modified classifier to get the new labelled datasets. The steps are repeated till convergence is attained.

Below is the pseudo-code for SM algorithm:

Pseudo-code for Variance-adjusted SVM:

1) Initialization phase:
   a) Name original training data labelled 1 and -1 as $\text{Train}_1$ and $\text{Train}_{-1}$ respectively.
   b) Calculate covariance matrices for $\text{Train}_1$ and $\text{Train}_{-1}$ - $S_{y=1}$ and $S_{y=-1}$ respectively.

2) Iteration phase:
   a) Perform Cholesky Decomposition of covariance matrices to obtain $C_{y=1}$ and $C_{y=-1}$, where $S_{y=1}=C_{y=1}C_{y=1}{}^T$ and $S_{y=1}=C_{y=-1}C_{y=-1}{}^T$.
   b) Calculate $C_{y=1}^{-1}\text{Train}_1$ and $C_{y=-1}^{-1}\text{Train}_{-1}$ to transform $\text{Train}_1$ and $\text{Train}_{-1}$ data from input space to Euclidean space.

c) Perform support vector classification on $\text{Train}_1$ and $\text{Train}_{-1}$ data in the Euclidean space and calculate the parameter vector $\theta_{\text{Euclidean}}$.
d) Perform linear SVM on the $\text{Train}_1$ and $\text{Train}_{-1}$ data (identified in initialization stage) in the input space and calculate the equation of the linear classifier $\Theta_{\text{Input}}^{\text{T}}\text{x}+\Theta_0=0$.
e) Adjust $\Theta_0$ to $\Theta_0'$ so that $\Theta_{\text{Input}}^{\text{T}}\text{x}+\Theta_0'=0$ divides the margin space in ratio $\sqrt{\frac{\theta_{Euclidean}^T C_{y=-1}^{-1} S_{y=-1}^{-1} C_{y=-1} \theta_{Euclidean}}{\theta_{Euclidean}^T C_{y=1}^{-1} S_{y=1}^{-1} C_{y=1} \theta_{Euclidean}}}$.
f) Label test datapoints as +1 and -1 based on $\Theta_{\text{Input}}^{\text{T}}\text{X}_{\text{Test}}+\Theta_0' \geq 0$ or $\leq 0$.
g) Add the test datapoints to $\text{Train}_1$ and $\text{Train}_{-1}$ based on their labels identified in step (f) to obtain new $\text{Train}_1$ and $\text{Train}_{-1}$ data.
h) Re-Calculate covariance matrices for new $\text{Train}_1$ and $\text{Train}_{-1}$ to get new $\text{S}_{\text{y}=1}$ and $\text{S}_{\text{y}=-1}$ respectively
i) Repeat steps (a) to (h) till convergence.

3) Convergence criteria:
a) Check if the test data assignments have stopped changing have stopped moving (i.e., the changes in test data labels are below a certain threshold).
b) If test data has converged, terminate the algorithm.

End Algorithm

## 4 Comparison with existing work and whitening algorithms

The CSVM model and algorithm proposed essentially transforms the data into Euclidean space by decorrelating and standardizing the data and then performs support vector classification in that space. This process is akin to data whitening. Data whitening has become a standard data pre-processing technique with many whitening algorithms being widely used e.g. PCA Whitening, ZCA Whitening etc. However, the novelty of this study is threefold:

- While data whitening leads to superior performance of support vector machines in many cases, the literature is divided on the reasons for this. In this study we provide a vector space explanation of why whitening works: ***whitening transforms the data from non-Euclidean space/input space to Euclidean space, and it is in Euclidean space that the equations for various ML models like support vector machines, KNN, K-Means clustering are based.***
- PCA/ZCA whitening algorithms perform whitening of the entire training data and then apply them on the test data. However, in many cases the data corresponding to various labels belongs to different populations and can have stark differences in their features. Hence whitening of the data needs to be carried out separately class-wise, as is done in this study.
- Instead of applying training data whitening transformation on test data, SM algorithm iteratively tries to identify labels of test data and thus calculate the population covariance matrix, leading to better performance vis-a-vis PCA and ZCA Whitening.

- ZCA or PCA whitening do not transform data to Euclidean space. On the other hand, Cholesky whitening is able to transform data to Euclidean space and it is the most suitable whitening for SVM (as seen in Results and Discussion).
- By starting from first principles and focusing on Mahalanobis Distance as a vector space transformation, we formulate the optimization problem in CSVM keeping it both vector space and dimensionally consistent, and thus address the limitations of previous studies done in variance adjusted SVM.

In addition, various studies have been carried out to incorporate data covariance via Mahalanobis distance in SVM optimization problem. The works of [13] [6] [9] have led to the following set of optimization equations:

$$\text{Min } \frac{1}{2}\theta^T\Sigma^{-1}\Theta + \lambda\sum_{i=1}^{N}\xi_i \quad (20)$$

$$\text{s.t. } y_i(\theta^T\Sigma^{-1}x_i+\theta_0) \geq 1\text{-}\xi_i, \quad \xi_i \geq 0 \qquad \forall i = 1,2,\ldots.N \quad (21)$$

while Huang et.al. [14] and Zafeiriou et al. [9] arrived at another set of equations:

$$\text{Min } \frac{1}{2}\theta^T\Sigma^{-1}\Theta + \lambda\sum_{i=1}^{N}\xi_i \quad (22)$$

$$\text{s.t. } y_i(\theta^T x_i+\theta_0) \geq 1\text{-}\xi_i, \ \xi_i \geq 0 \quad \forall i = 1,2,\ldots.N \quad (23)$$

with Zafeiriou et. al using scatter matrix instead of covariance matrix. While equations (25) to (28) have tried to incorporate the data covariance into optimization problem of SVMs with significant improvement in performance w.r.t. linear SVMs, they suffer from the following limitations:

a) It is not clear how they chose equations (20), (21) or (22). The derivation of variance adjusted objective function and constraints based on first principles are not clear in the studies;
b) There is vector space inconsistency in equations (22) and (23) i.e. equation (22) is in Euclidean space while (23) is in non-Euclidean space.
c) There is dimensional inconsistency in the constraint equation (21). As per Ke et al. [6], Mahalanobis distance is Euclidean distance after decorrelation and standardization of every random variable. If we consider the special case where correlation between variables is 0, then each variable is being standardized by the formula $\frac{X}{\sigma^2}$. However, in univariate setting, standardization of random variables is carried out by using the formula $\frac{(X)}{\sigma}$. Hence in multivariate setting the standardization the random variables need to be carried out using $\Sigma^{\frac{-1}{2}}$, not $\Sigma^{-1}$. This is leading to dimensional inconsistency.

In this study, by starting from first principles and focusing on Mahalanobis Distance as a vector space transformation, we formulate the optimization problem in CSVM keeping it both vector space and dimensionally consistent, and thus address the limitations of previous studies.

## 5 Results and Discussion

For experimental verification of covariance-adjusted SVC (CSVC) as well as the effectiveness of SM algorithm, the CSVC model was applied on five popular datasets- Breast Cancer Wisconsin Dataset, OSHA Dataset, Diabetes Dataset, Red Wine Dataset and Pulsar Dataset. The objective of selecting the 5 datasets was to check for generalizability of the model in various domains- healthcare, astronomy, quality, and safety/text mining domains. First the dataset was split into training and validation data in the ratio 80:20. Then CSVC model was applied on the data and the classification table giving the accuracy, precision, recall and F1 scores and Receiver Operating Characteristic (ROC) curves giving the Area Under Curve (AUC) were obtained. In addition, the effectiveness of CSVC model is compared with other widely used support vector classification models by applying various support vector kernels - linear, RBF, Sigmoid and Polynomial- on the datasets and obtaining the classification tables and ROC curves. The accuracy, precision, recall, F1 scores and AUC values were compared.

In addition, owing to similarity with whitening algorithms, we compared CSVM with PCA and ZCA whitening cum SVM approach- PCA and ZCA whitening were carried out on the 5 datasets and then linear SVM was used and classification table and ROC curves were obtained. The accuracy, precision, recall, F1 scores and AUC values were then compared for our model, PCA whitening-linear SVC and ZCA whitening-linear SVC.

The accuracy, precision, recall and F1 scores of the SVM kernels and whitening models over the 5 datasets is given in Table 1, 2, 3 and 4 respectively. It can be seen that our model CSVM-Cholesky Decomposition attains highest accuracy and F1 scores for all 5 datasets. CSVM Model has highest precision and recall values for 4 out of 5 datasets and for the $5^{th}$ dataset it has the second highest precision and recall values.

Table 1: Comparison of accuracy of SVM kernels and whitening models over 5 datasets

| | Breast Cancer | OSHA | Diabetes | Red Wine | Pulsar |
|---|---|---|---|---|---|
| CSVM- Cholesky | **0.974** | **0.771** | **0.779** | **0.753** | **0.981** |
| SVM- Linear | 0.956 | 0.768 | 0.760 | 0.731 | 0.979 |
| SVM-RBF | 0.947 | 0.769 | 0.727 | 0.637 | 0.973 |
| SVM- Sigmoid | 0.465 | 0.702 | 0.656 | 0.425 | 0.925 |
| SVM- Poly | 0.947 | 0.752 | 0.747 | 0.644 | 0.971 |
| SVM Linear PCA | 0.947 | 0.768 | 0.760 | 0.728 | 0.979 |
| SVM Linear ZCA | 0.939 | 0.768 | 0.760 | 0.744 | 0.979 |

Table 2: Comparison of precision of SVM kernels and whitening models over 5 datasets

| | Breast Cancer | OSHA | Diabetes | Red Wine | Pulsar |
|---|---|---|---|---|---|
| CSVM- Cholesky | **0.974** | **0.770** | **0.762** | **0.750** | 0.959 |
| SVM- Linear | 0.960 | 0.767 | 0.738 | 0.733 | **0.962** |
| SVM-RBF | 0.961 | 0.772 | 0.702 | 0.651 | 0.954 |
| SVM- Sigmoid | 0.380 | 0.688 | 0.631 | 0.420 | 0.770 |
| SVM- Poly | 0.961 | 0.753 | 0.738 | 0.709 | 0.939 |
| SVM Linear PCA | 0.948 | 0.767 | 0.738 | 0.731 | **0.962** |
| SVM Linear ZCA | 0.937 | 0.767 | 0.738 | 0.745 | **0.962** |

Table 3: Comparison of recall of SVM kernels and whitening models over 5 datasets

| | Breast Cancer | OSHA | Diabetes | Red Wine | Pulsar |
|---|---|---|---|---|---|
| CSVM- Cholesky | **0.970** | **0.742** | **0.747** | 0.747 | **0.924** |
| SVM- Linear | 0.946 | 0.738 | 0.736 | 0.736 | 0.906 |
| SVM-RBF | 0.930 | 0.736 | 0.691 | 0.606 | 0.876 |
| SVM- Sigmoid | 0.401 | 0.686 | 0.635 | 0.419 | 0.762 |
| SVM- Poly | 0.930 | 0.716 | 0.686 | 0.603 | 0.876 |
| SVM Linear PCA | 0.939 | 0.738 | 0.736 | 0.734 | 0.906 |
| SVM Linear ZCA | 0.932 | 0.738 | 0.736 | **0.748** | 0.906 |

Table 4: Comparison of F1 scores of SVM kernels & whitening models over 5 datasets

| | Breast Cancer | OSHA | Diabetes | Red Wine | Pulsar |
|---|---|---|---|---|---|
| CSVM- Cholesky | **0.974** | **0.749** | **0.753** | **0.748** | **0.940** |
| SVM- Linear | 0.956 | 0.745 | 0.737 | 0.731 | 0.932 |
| SVM-RBF | 0.947 | 0.744 | 0.695 | 0.591 | 0.910 |
| SVM- Sigmoid | 0.465 | 0.687 | 0.632 | 0.419 | 0.766 |
| SVM- Poly | 0.947 | 0.723 | 0.696 | 0.568 | 0.904 |
| SVM Linear PCA | 0.947 | 0.745 | 0.737 | 0.728 | 0.932 |
| SVM Linear ZCA | 0.939 | 0.745 | 0.737 | 0.743 | 0.932 |

Finally, the performance comparison of the SVM kernels and PCA/ZCA whitening methods is compared by drawing the ROC curves and measuring the Area Under Curve (AUC) values. It can be seen that for all the 5 datasets, our CSVM Cholesky transformation model is having the highest AUC values. For OSHA data, CSVM AUC is joint highest with Linear, RBF, PCA and ZCA Whitening and for red wine data, CSVM is joint highest with ZCA Whitening AUC.

Hence it can be seen that transductive CSVM is out-performing other inductive SVMs. But how does it perform against other transductive SVM models? For that, we compare transductive CSVM against standard transductive SVM of scikit-learn and measured the accuracies for the datasets. The results are presented in Table 5:

Table 5: Comparison of accuracy of CSVM and transductive SVM over 5 datasets

| | OSHA | Breast | Red Wine | Pulsar | Diabetes |
|---|---|---|---|---|---|
| CSVM | **0.771** | **0.974** | **0.753** | **0.981** | **0.779** |
| TSVM | 0.769 | 0.956 | 0.734 | 0.979 | 0.76 |

Hence it can be seen transductive covariance-adjusted CSVM model is outperforming standard transductive TSVM.

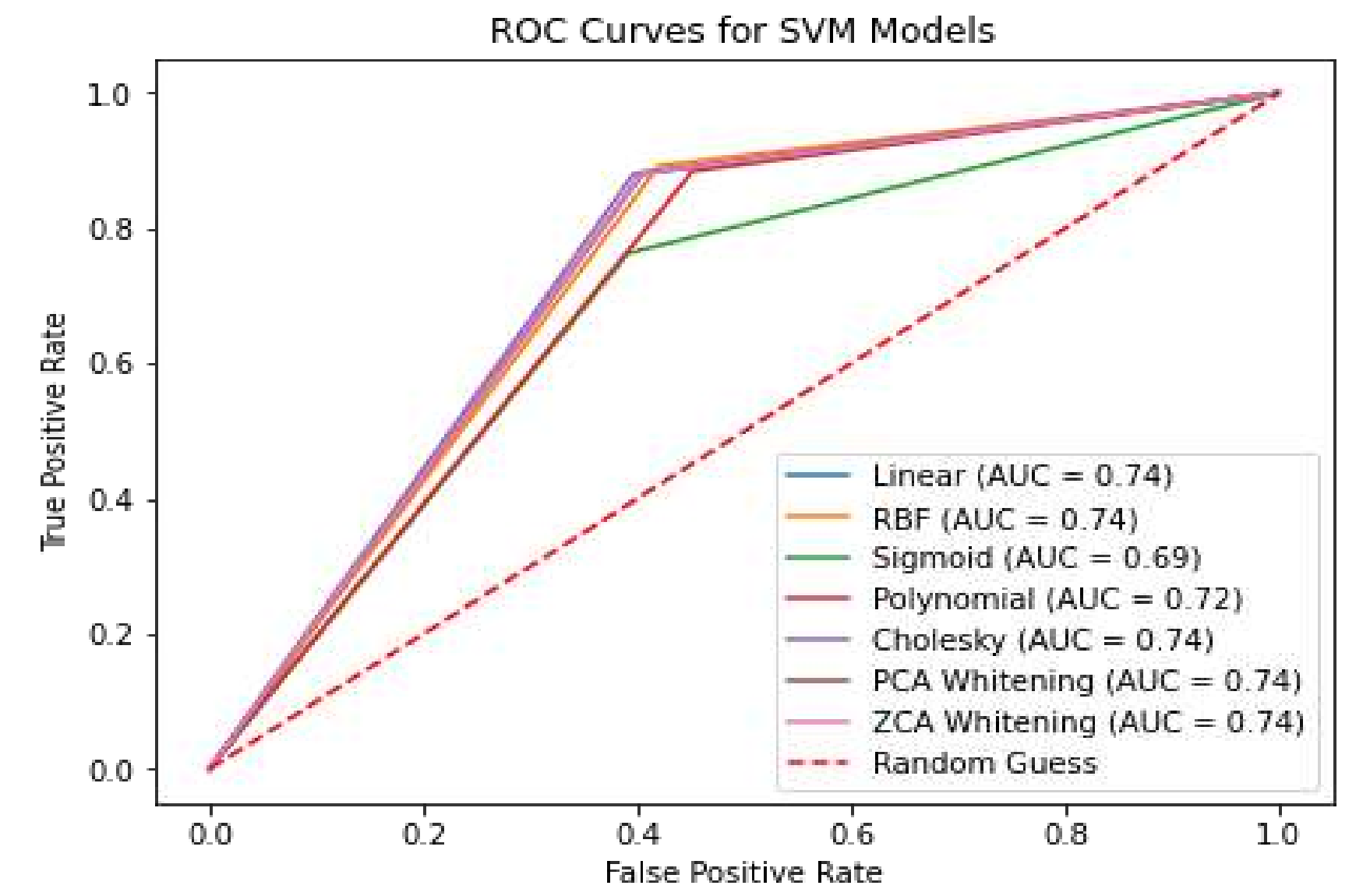

Fig 2: ROC Curves and AUC values for SVM kernels and whitening models for OSHA Data

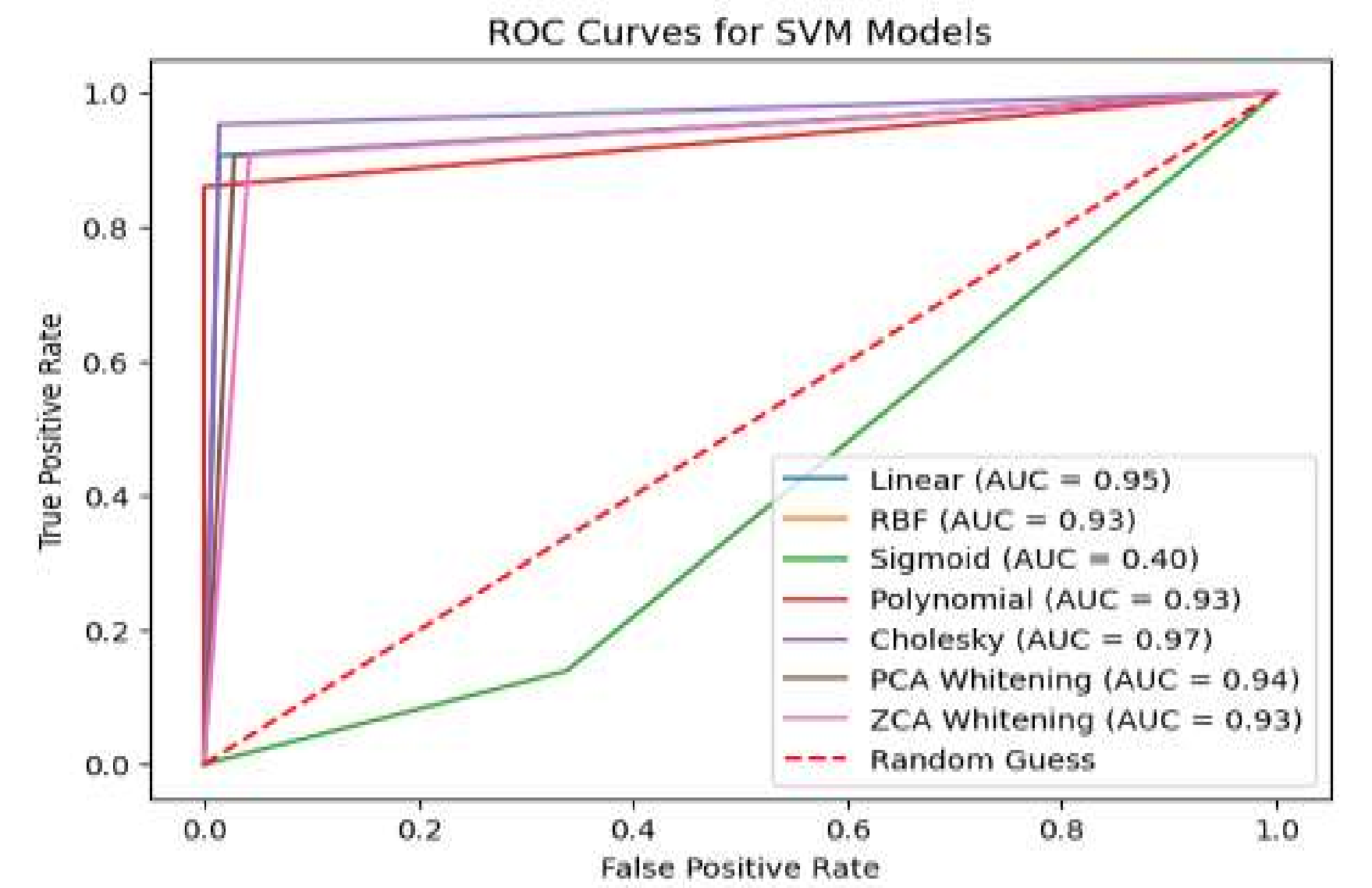

Fig 3: ROC Curves and AUC values for SVM kernels and whitening models for Breast Cancer data

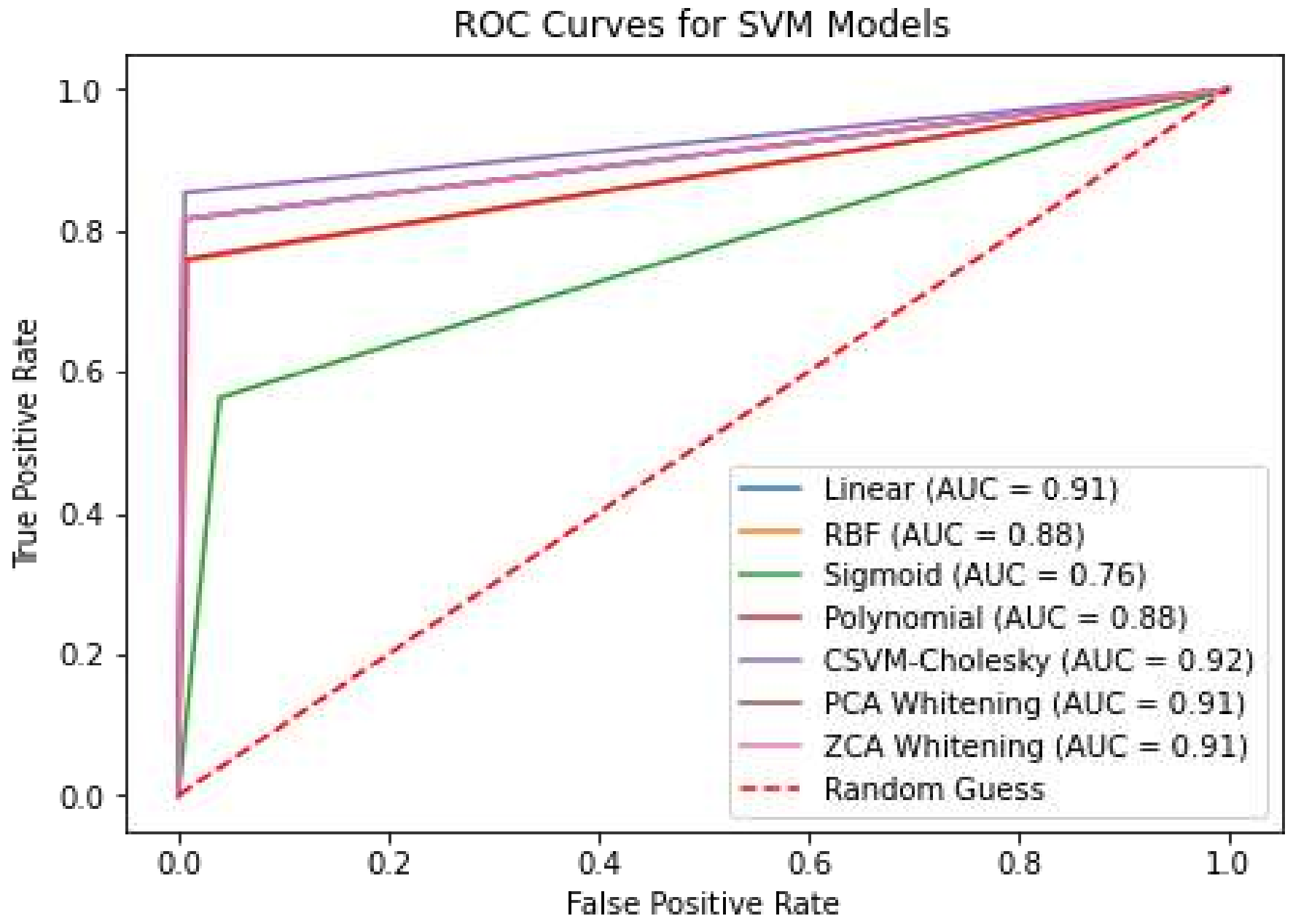

Fig 4: ROC Curves and AUC values for SVM kernels and whitening models for Pulsar data

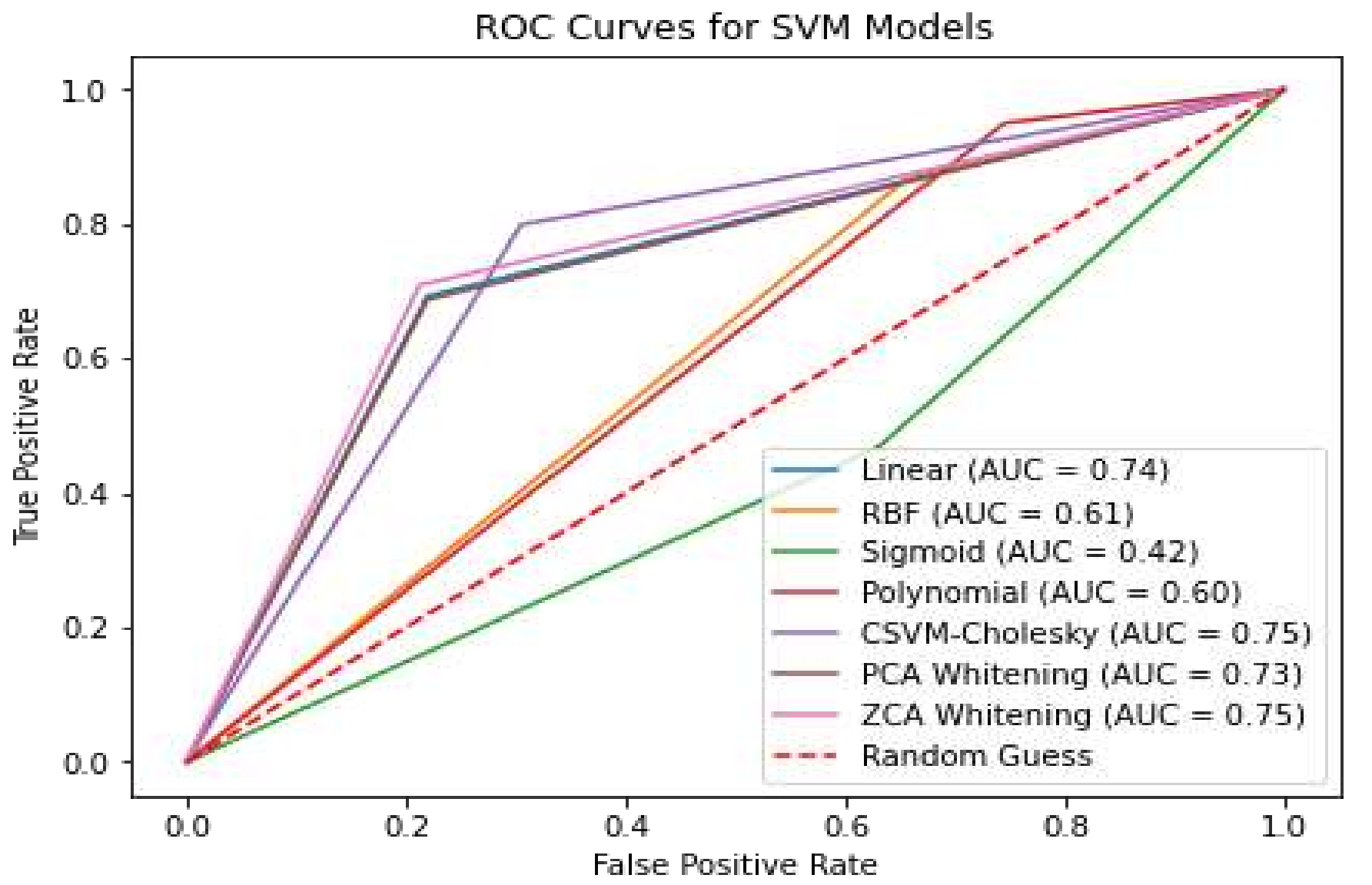

Fig 5: ROC Curves and AUC values for SVM kernels and whitening models for Red wine data

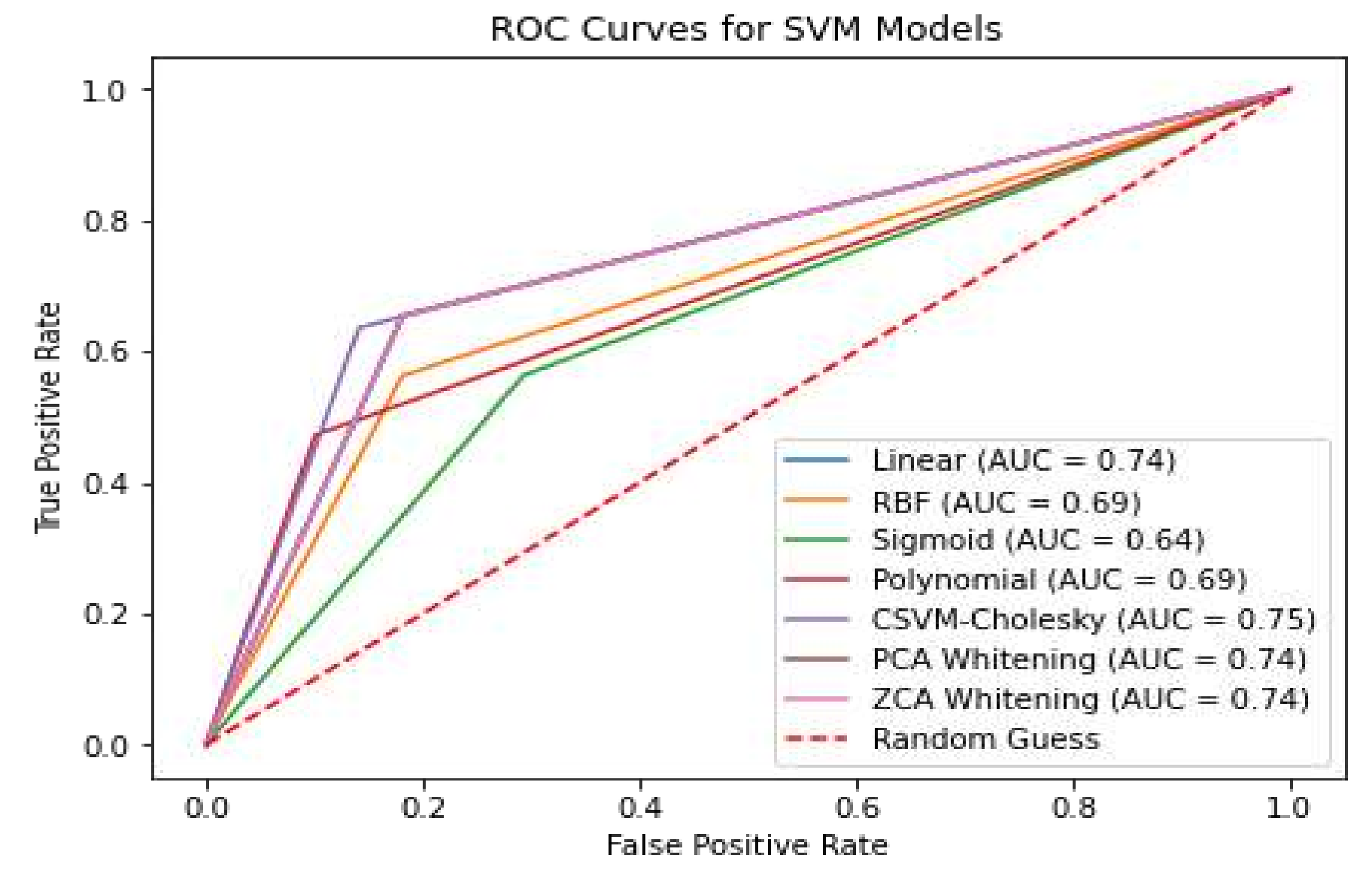

Fig 6: ROC curves and AUC values for SVM kernels and whitening models for diabetes data

This the values of accuracy, precision and F1 scores as well as the area under curve for the ROC curves provide experimental verification for the CSVM model and show that lemmas 3.1, 3.2 and 3.3 are valid.

## 6 Conclusion, limitations, and future work

It can be seen that the CSVM model is a powerful tool that gives better results compared to traditional SVM kernels. The purpose of experimenting and comparing with traditional SVM approaches over 5 datasets was to validate the findings of lemma 3.1, 3.2 and 3.3. Hence in this study we demonstrate that (1) SVM optimization problem formulation and solutions are optimal in the Euclidean space, and formulating and solving them in the input space carries risk of misclassification; (2) A binary class problem requires one classifier in the Euclidean space but two classifiers in the Non-Euclidean space, while an N class problem has one classifier in Euclidean Space but N classifiers in the Non-Euclidean input space; (3) The distance of the classifiers from their respective margin boundaries is a function of their intra-class covariances.

Despite the performance of Cholesky transformation-CSVM, it suffers from certain drawbacks. First, it requires the knowledge of the population covariance structure of the data distribution. In absence of information about population covariance, the transformation of data to Euclidean space becomes difficult. While SM algorithm has been proposed to address this limitation, it is a heuristic algorithm and though it gives the best classification performance, perfect classification is yet to be achieved. Secondly, the computational complexity of Cholesky kernel is higher than traditional linear SVM as extra steps of calculating covariance matrices and Cholesky decomposition are involved. Hence it leads to the following dilemma: is the increase in classification performance worth the computational complexity?

Secondly, for some datasets it was observed that substituting $\sqrt{\frac{\theta_{Euclidean}^{T} C_{y=-1}^{-1} S_{y=-1}^{-1} C_{y=-1} \theta_{Euclidean}}{\theta_{Euclidean}^{T} C_{y=1}^{-1} S_{y=1}^{-1} C_{y=1} \theta_{Euclidean}}}$ with $\frac{\theta_{Euclidean}^{T} C_{y=-1}^{-1} S_{y=-1}^{-1} C_{y=-1} \theta_{Euclidean}}{\theta_{Euclidean}^{T} C_{y=1}^{-1} S_{y=1}^{-1} C_{y=1} \theta_{Euclidean}}$ yielded higher accuracy, precision, recall and F1 scores. It is yet to be understood why this is so.

Considering the limitations, future work will involve finetuning the SM algorithm so that we can reduce the computational complexity or achieve classification as close to 100% as possible and to understand the reason for better classification performance when we use $\frac{\theta_{Euclidean}^{T} C_{y=-1}^{-1} S_{y=-1}^{-1} C_{y=-1} \theta_{Euclidean}}{\theta_{Euclidean}^{T} C_{y=1}^{-1} S_{y=1}^{-1} C_{y=1} \theta_{Euclidean}}$.

**Disclosure of Interests.** The authors declare that they have no known competing financial interests or personal relationships that could have appeared to influence the work reported in this paper.